\documentclass[11pt, letterpaper]{article}
\usepackage{latexsym}
\usepackage{amssymb}
\usepackage{graphicx}
\usepackage{amsmath}

\usepackage[english]{babel}

\title{Formal semantics of  language and the Richard-Berry paradox}
\author{Stefano Crespi Reghizzi\\Politecnico di Milano\\
DEI - Dipartimento di Elettronica e Informazione\\
Piazza Leonardo da Vinci, 32 Milano 20133\\
Tel. +39 0223993518\\
stefano.crespireghizzi@polimi.it}

\begin{document}
\maketitle
 \abstract{The classical logical antinomy known as Richard-Berry paradox is combined with plausible assumptions about the size i.e. the descriptional complexity, of Turing machines formalizing certain sentences, to show that formalization of language leads to contradiction.
\\}
\\
\vspace{20pt}
\\
\vspace{20pt} \emph{Article type}: letter
\\

Formal semantics of language has roots in the disciplines of logic, philosophy and mathematics with pioneering work by Frege and Russell in the early 20th century and important addition by Tarski and Carnap in particular. It emerged as an important area of linguistics in the 1970s from seminal research by Richard Montague.
 In recent years fruitful applications of  aspects of formal semantics have been made in computational
 linguistic, artificial intelligence, and natural language processing.
 \\
 Various fragments of English and other languages have been
 carefully formalized, but the ultimate objective of complete formalization
 is still remote, and not everybody, to say the least, believes it to
 be realistic.
 \\
This correspondence exhibits an internal contradiction inherent in language formalization, in the classical form of a logical paradox.
\paragraph{The Richard-Berry paradox.}
It is good to start by recalling the logical paradox described in  the \emph{Principia Mathematica} (Russell and Whitehead, 1917), where the authors say the paradox ``was suggested to us by Mr. G.G. Berry of the Bodleian Library''. The Richard-Berry Paradox is the definition of a number as
\begin{equation}\label{R-B}
   \text{``the
least number that cannot be defined in fewer than twenty words.''}
\end{equation}
The antinomy is explained by (Li and Vit{\'a}nyi 1997) as follows:
\begin{quotation}
If this number exists, we have just described it in thirteen words, contradicting its definitional statement. If such number does not exist, then all natural numbers can be described in fewer than twenty words.
\\
\ldots
\\
Formalizing the notion of ``definition'' as the shortest program from which a number can be computed by the reference [Turing] machine $U$, it turns out that the quoted statement (reformulated appropriately) is not an \emph{effective description}.
\end{quotation}
It is known that the proof of the impossibility of calculating the number described in (\ref{R-B}) gives a way of rephrasing G{\"o}del's incompleteness theorem.
\par We observe in statement (\ref{R-B}) the word ``least'' has
the usual arithmetic sense, as well as the meaning that the number referred to is the \emph{first} encountered in the enumeration $1,2, 3, \ldots $. The paradox still holds if we replace the ``least number'' by the ``first number'' (in the enumeration) or,
 if, instead of natural numbers, we use other
discrete enumerable structures, such as rational numbers. Would the paradox still hold if numbers are replaced by texts?
\par
Moving from this, the present note argues that the statement that natural language can be completely formalized leads to antinomy.
\paragraph{The formal semantics paradox.}
Let us focus on  some natural language, say English, and assume that any
  text can be precisely formalized by a Turing
machine (or for that matter by any other computationally complete formalism). This means that given a text, a procedure exists  translating it to a formal definition, as the description or program of a Turing machine. We do not rule out the possibility that a text be formalized in different ways, corresponding to different Turing machines.
\\
A Turing machine description can be encoded into a binary string, and the strings describing Turing machines can be enumerated by increasing \emph{size}. Thus any machine has a position in the enumeration, and it makes sense to say that a machine comes before or after another. The notion of machine size can be made rigorous enough, as done in the theory of complexity of Kolmogorov, Chaitin, and Solomonoff, for which we refer to the classical book (Li and Vit{\'a}nyi 1997). Here we take size as synonymy of the position of a machine in the above enumeration.
\\
Since a text may have more than one formal definition, we consider the first one in the enumeration as the reference definition. Thus the reference machine is the one of least size among the definitions of a text.\\
The size of the reference machine formalizing a certain text will be called the \emph{formal complexity} of the text.
\\
Now we can imagine to sort the English texts in ascending order of their formal complexity. This means text one precedes text two, if their respective formal definitions as Turing machines, which we have assumed to be computable, are in that order in the enumeration.
\par
Since texts are now ordered, it makes sense to consider
\begin{equation}\label{SmallestTextSuch}
   \text{``the first text such  that its formal
 complexity is not less than twenty.''}
\end{equation}
Sentence (\ref{SmallestTextSuch})  will be denoted by $t(20)$, to emphasize that it is parameterized by the number twenty. By changing the numerical parameter, we may obtain similar sentences denoted as $t(21)$, and so on.
\par Before we proceed with the main argument, we have to make
explicit two intuitively reasonable assumptions on the complexity of formal descriptions.
\begin{description}
    \item[Unboundedness assumption. ] For each integer $n$, there
    exists a text having formal complexity greater than $n$.
    \\
    The idea is that texts may require arbitrarily complex formal descriptions.
    \item[Logarithmic complexity. ] Consider the family
    of sentences $ t(20),t(21), \ldots$. For any integer $n$ greater than twenty, the formal
    complexity of $t(n)$ does not exceed the formal complexity of
    $t(20)$ by more than a quantity proportional to $log(n)$.
    \\
   To justify the assumption, consider that the formal description of $t(n)$ includes two parts:
   one is independent of $n$ and therefore has a size less than
   the size of $t(20)$; the other part has a logarithmic
   complexity, since it is well known that integers can be encoded by a
    positional number representation having a logarithmic number of digits.
\end{description}
\paragraph{Main argument.}
Two cases are possible.
\begin{enumerate}
    \item First suppose ``the first text'' referred to in
(\ref{SmallestTextSuch}) exists, and consider the complexity of the formal description of sentence $t(20)$. Two subcases are possible.
\begin{enumerate}
    \item The formal complexity of $t(20)$ is less than twenty. Since $t(20)$ provides a definition of
    `the first text\ldots'',  we
    have found a formal description of it
    contradicting the definition.
    \item The formal complexity of $t(20)$ is $k\ge 20$. Then,
   from the logarithmic complexity assumption, one can find a sufficiently large  integer $K$  greater than $k$ such that the
   formal complexity of sentence $t(K)$ is less than $K$, thus
   obtaining a contradiction for the ``the first text such that its formal
 complexity is not less than $K$.''
\end{enumerate}
    \item Second, suppose  ``the first text'' referred to in
(\ref{SmallestTextSuch}) does not exists. Then any text would have formal complexity less than twenty, contradicting the unboundedness assumption.
\end{enumerate}
To conclude, we observe the classical Richard-Berry paradox relies on an enumeration of English sentences ordered by the number of words they contain, i.e. essentially by their length. This version of the paradox orders the sentences according to the length of their formalizations (say by Turing machines), assumed to exist.  The ensuing contradiction proves that a complete computational formalization of natural language sentences is impossible.
\paragraph{References\\}
\noindent Li Ming and Vitanyi P. (1997). An Introduction to Kolmogorov Complexity and its Applications. (New York: Springer).
\\
Russell B. and  Whitehead A.N. (1917). Principia Mathematica. (Cambridge: University Press).
\end{document}